\documentclass[letterpaper]{article} 
\usepackage{aaai25}  
\usepackage{times}  
\usepackage{helvet}  
\usepackage{courier}  
\usepackage[hyphens]{url}  
\usepackage{graphicx} 
\usepackage{natbib}  
\usepackage{caption} 
\usepackage{algorithm}
\usepackage{algorithmic}
\usepackage{amsmath}
\usepackage{booktabs}
\usepackage{xcolor}
\usepackage{marvosym}

\usepackage{newfloat}
\usepackage{listings}
\DeclareCaptionStyle{ruled}{labelfont=normalfont,labelsep=colon,strut=off} 
\floatstyle{ruled}
\newfloat{listing}{tb}{lst}{}
\floatname{listing}{Listing}
\title{Recoverable Compression: A Multimodal Vision Token Recovery Mechanism Guided by Text Information}
\author{
    Yi Chen\textsuperscript{\rm 1,2}, Jian Xu\textsuperscript{\rm 1,2}, Xu-Yao Zhang\textsuperscript{\rm 1,2}, Wen-Zhuo Liu\textsuperscript{\rm 1,2}, Yang-Yang Liu\textsuperscript{\rm 1,2}, Cheng-Lin Liu\textsuperscript{\rm 1,2}\textsuperscript{(\Letter)}\\
}
\affiliations{
    \textsuperscript{\rm 1}School of Artificial Intelligence, \\
    University of Chinese Academy of Sciences, Beijing 100049, China\\
    \textsuperscript{\rm 2}State Key Laboratory of Multimodal Artificial Intelligence Systems (MAIS),\\ Institution of Automation, Chinese Academy of Sciences, Beijing 100190, China\\
   {\{yi.chen, xyz, liucl\}@nlpr.ia.ac.cn}, {\{jian.xu, liuwenzhuo2020, liuyangyang2021\}@ia.ac.cn}
}

\usepackage{bibentry}

\begin{document}

\maketitle

\begin{abstract}
With the advancement of large-scale language modeling techniques, large multimodal models combining visual encoders with large language models have demonstrated exceptional performance in various visual tasks. Most of the current large multimodal models achieve this by mapping visual features obtained from the visual encoder into a large language model and using them as inputs alongside text for downstream tasks. Therefore, the number of visual tokens directly affects the training and inference speed of the model. There has been significant work on token pruning for visual transformers, but for large multimodal models, only relying on visual information for token pruning or compression may lead to significant loss of important information. On the other hand, the textual input in the form of a question may contain valuable information that can aid in answering the question, providing additional knowledge to the model. To address the potential oversimplification and excessive pruning that can occur with most purely visual token pruning methods, we propose a text information-guided dynamic visual token recovery mechanism that does not require training. This mechanism leverages the similarity between the question text and visual tokens to recover visually meaningful tokens with important text information while merging other less important tokens, to achieve efficient computation for large multimodal models. Experimental results demonstrate that our proposed method achieves comparable performance to the original approach while compressing the visual tokens to an average of 10\% of the original quantity.
\end{abstract}

%
\begin{links}
    \link{Code}{https://github.com/banjiuyufen/Recoverable-Compression}
\end{links}

\section{Introduction}

With the continuous development of deep learning and semiconductor technology, large language models (LLMs) \cite{brown2020language,ouyang2022training,jiang2023mistral,touvron2023llama} have made amazing achievements in natural processing language tasks. LLMs usually adopt a Transformer structure with hundreds of billions of parameters and use large-scale text language materials for pre-training. By scaling up data size and model parameters, the LLMs can better understand natural language and generate high-quality text based on the given context.

The multimodal large language models (MM-LLMs) \cite{liu2023improved,liu2024llava,openai2023gpt, team2023gemini} take LLMs as the core and use this powerful language generation, zero-shot transfer, and context learning capabilities to solve multi-modal tasks. Specifically, MM-LLMs use a vision encoder, such as the NFNet-F6 \cite{brock2021high}, Vision Transformer (ViT) \cite{dosovitskiy2020image} and the CLIP \cite{radford2021learning}, to convert them into vision features, align them with the text input of LLMs through the projection layer, and finally concatenate the vision features and text features into the LLMs.

\begin{figure}[!htb]
\centering
\includegraphics[width=1.\linewidth]{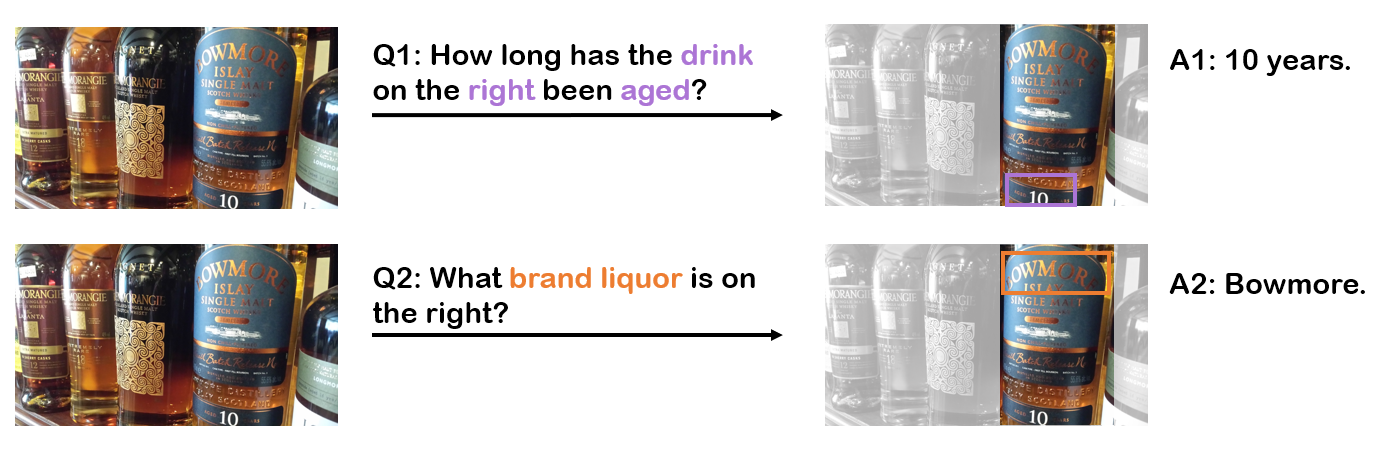}
\caption{The key areas of the same image under different questions.} 
\label{fig: text_inf}
\end{figure}  

Due to the massive number of model parameters, both training and inference of MM-LLM require significant computational resources. While the vision encoder of MM-LLM has relatively fewer parameters compared to the entire LLM, subsequent LLMs contribute significantly to the computational demands. Therefore, there are two main approaches to improve the efficiency of LLM. The first approach is to directly use LLMs with smaller parameter sizes. By reducing the number of parameters in the model, computational requirements can be significantly reduced, leading to faster training and inference. The other approach is to reduce the output of the vision encoder. Since LLMs typically employ Transformer structures, the computational cost of Transformers often grows quadratically with the length of the input context. By reducing the input length of LLMs, the overall training and inference speed of the model can be greatly improved. Both approaches offer potential solutions for enhancing the efficiency of MM-LLM, allowing for faster and more resource-efficient training and inference.

Previous research efforts have focused on the first method, which achieves a smaller number of parameters by replacing the LLM backbone, such as Chu et al \cite{chu2023mobilevlm}. implemented MM-LLM for mobile devices by using LLM backbones with $1.4B$ and $2.7B$ parameters. Yuan et al \cite{yuan2023tinygpt}. proposed TinyGPT-V, a new multimodal large language based on small backbones. which can be suitable for local deployment and inference tasks on various devices with 8G graphics memory with the $2.8B$ parameter. However, as the LLM backbone becomes smaller, the reasoning ability of LMM is sacrificed, resulting in a decline in various performance results.

Recently, related works have used pruning methods to achieve efficient inference of MM-LLM. For example, Wang et al \cite{wang2023smarttrim}. integrated lightweight modules into the original backbone to identify and remove redundant tokens and attention heads in each layer to accelerate the training and inference process of the model. Shang et al \cite{shang2024llava}. proposed an adaptive token pruning strategy to reduce the number of vision tokens through clustering.

Given that MM-LLM often employ ViT structures in their vision encoders, many pruning methods developed for ViT models are also applicable to MM-LLM based on ViT encoders. However, relying solely on individual modality information during pruning can result in the loss of important information in multimodal tasks and models. Additionally, the textual modality contains valuable information that can enhance the knowledge and assist in addressing questions effectively \cite{ganz2024question}. Such as Figure\ref{fig: text_inf}, even for the same image, different questions correspond to different regions of detail. Therefore, exploring how to combine vision and textual modality information for pruning and achieving efficient training and inference of MM-LLM is a valuable and relevant research and application topic.

To achieve this, we propose a training-free semantic-guided dynamic visual token recovery mechanism. Specifically, we compute the similarity between visual tokens and the question text, which serves as the basis for subsequent token recovery guided by the question text. As the class token in the ViT represents the global representation of the image, we calculate the similarity between the class token and other visual tokens as a criterion to perform initial token filtering. Next, we reclaim visual tokens with high text similarity from the remaining visual tokens. Finally, we merge the remaining unimportant tokens. In the above steps, the dynamic scale filtering method is used to filter out important tokens.
We conducted experiments on multiple MM-LLM evaluation datasets, and the results show that our proposed method can achieve token compression to around $10\%$ of the original quantity while maintaining competitive performance compared to the original model.

Our main contributions are summarized as follows:
\begin{itemize}
    \item We propose a multimodal large language model token recovery mechanism. Unlike other pruning methods, our approach goes beyond pruning and incorporates a secondary recovery of the remaining tokens to ensure that important information is preserved as much as possible. 
    \item By combining both modalities in the pruning process, our method dynamically filters vision tokens that are crucial for both modalities, enabling efficient inference in MM-LLM. We achieved an average token compression rate of $9$x on multiple datasets while maintaining competitive performance.
    \item Our training-free method offers simplicity and efficiency. It can be easily implemented without the need for additional training. 
\end{itemize}

\section{Related Work}
\subsection{Vision Token Compression}
Most MM-LLMs utilize a ViT-based vision encoder and a Transformer Decoder-based LLMs. For the Transformer, the computational complexity increases quadratically with the token length in the self-attention layers. Therefore, by reducing the number of tokens obtained from the vision encoder of MM-LLM, the computational efficiency of MM-LLM can be significantly improved.

Currently, many researchers are working on achieving efficient ViT models, and token compression has become one of the main research directions. Token compression can be divided into token pruning and token merging strategies. Token pruning involves evaluating the importance of different tokens based on defined criteria, preserving important tokens, and discarding unimportant ones. For example, Rao et al. \cite{rao2021dynamicvit} proposed a dynamic pruning method that prunes redundant tokens gradually and dynamically at each layer of the model based on the sparsity of visual attention and estimates the importance scores of each token using current features. Kong \cite{kong2022spvit} and Xu \cite{xu2023no} went even further, suggesting that unimportant tokens, should not be simply discarded but rather integrated or further manipulated to avoid the problem of permanent loss of image information caused by improper pruning.

Token merging combines similar tokens to discard unimportant background tokens and achieve efficient token compression by merging foreground tokens. Chen et al. \cite{chen2023diffrate} associated the loss function with the compression rate to automatically learn different token compression rates for different layers, combining pruning and merging simultaneously. To ensure the reliability of the merging process, Long et al. \cite{long2023beyond} considered both token importance and diversity for pruning and further merged similar tokens. Similarly, Lee et al. \cite{lee2024multi} also emphasized the need to consider diverse relationships between tokens during token merging. They designed multiple criteria to gradually fuse tokens, achieving the optimal balance between speed and accuracy.

For MM-LLMs, if we use methods designed specifically for ViTs that only consider visual modality for token compression, some tokens containing important information may be lost during the compression process, leading to a decrease in performance.

\begin{figure*}[!htb]
\centering
\includegraphics[width=0.8\linewidth]{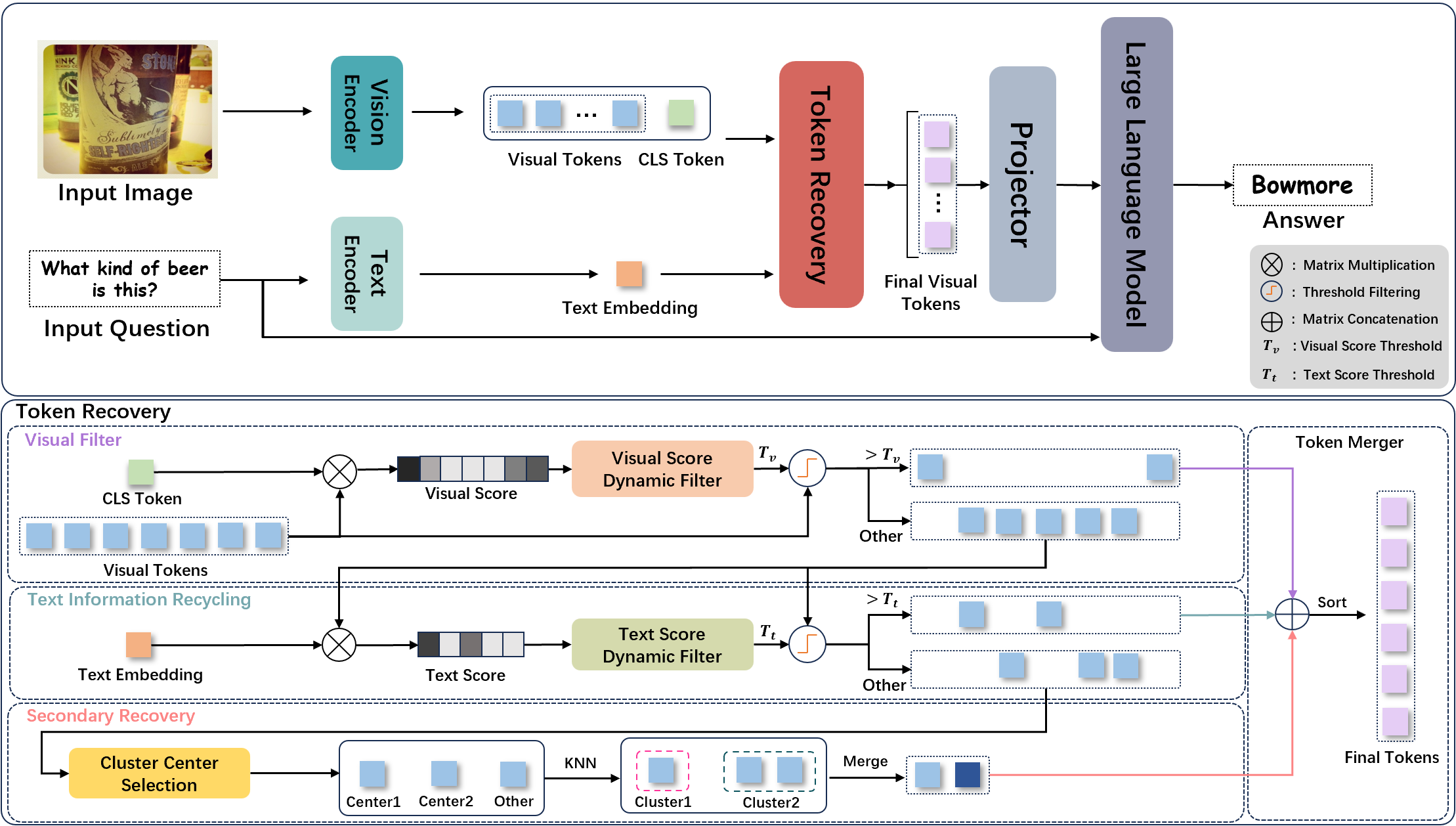}
\caption{Overview of the proposed multimodal vision token recovery mechanism guided by text information framework. The lower part shows the detailed framework of our proposed recovery mechanism.} 
\label{fig: overview}
\end{figure*}  

\subsection{Enhancing MM-LLM with Text Information}
Recently, research has focused on the beneficial impact of textual information on MM-LLM. For instance, Ganz et al. \cite{ganz2024question} proposed a question-aware visual Transformer for multimodal reasoning, which directly embeds question awareness into the visual encoder, allowing visual features to pay more attention to image details relevant to the posed questions. Cao et al. \cite{cao2024madtp} discovered that each modality has visual tokens that are important for their respective modalities. Therefore, they proposed a modality alignment-guided dynamic token pruning method to ensure that pruned tokens are not important to any modality. Liu et al. \cite{liu2024hrvda} introduced content filtering mechanisms and instruction filtering modules, which filter out visually irrelevant tokens and instruction-agnostic tokens respectively, thereby enabling efficient model training and inference for high-resolution images. Shi et al. \cite{shi2023crossget} proposed a token cross guidance mechanism for accelerating visual language transformer, which combines tokens adaptively in real-time during the inference process, significantly reducing computational costs while maintaining high performance.

Based on the above work, we propose a concept parallel to pruning and merging, called the token recovery mechanism. This mechanism gets information from text modalities to sort tags that have been pruned and discarded by visual modalities. Then restore tokens with high semantic similarity, ensuring that important semantic information can be retained even after pruning. For the tokens filtered out in the second round, we use KNN to merge them and add them back to the previously selected tokens. This ensures that important information in the background is not discarded.

\section{Method}
\subsection{Overview}
To minimize the loss of important information during the token compression process, we propose a text information-guided dynamic visual token recovery mechanism. The framework of this method is illustrated in Figure \ref{fig: overview}. Firstly, the image and the question are separately encoded by visual and text encoders, resulting in visual tokens and text embeddings. Then, these outputs are fed into the token recovery module, which consists of four steps:

\begin{enumerate}
\item \textbf{Visual Filter\quad} Calculate the similarity between the visual class token and other visual tokens, generating visual scores. A dynamic scale filter algorithm is used to determine the threshold for the visual scores, and the top-k tokens based on the threshold are selected as the visual tokens with high scores.

\item \textbf{Text Information Recovery\quad} Calculate the similarity between the remaining tokens and the text embedding, generating text scores. Similarly, use a dynamic scale filter algorithm to determine the threshold for the text scores, and select the top-k tokens based on the threshold as the text tokens with high scores. This completes the first round of semantic-guided dynamic recovery.

\item \textbf{Secondary Recovery\quad} For the remaining tokens, apply the KNN to perform clustering and merge each cluster into a single token.

\item \textbf{Token Merger\quad} Concatenate all the tokens obtained from Steps 1, 2, and 3. It is worth noting that during the training phase, LLMs are trained on input sequences arranged according to the original token order. As a result, the input to LLM is highly sensitive to the sequence order. It is important to note that when merging tokens from Steps 1 and 2, the original order of tokens should be maintained.
\end{enumerate}

\begin{figure*}
\centering
\includegraphics[width=0.75\linewidth]{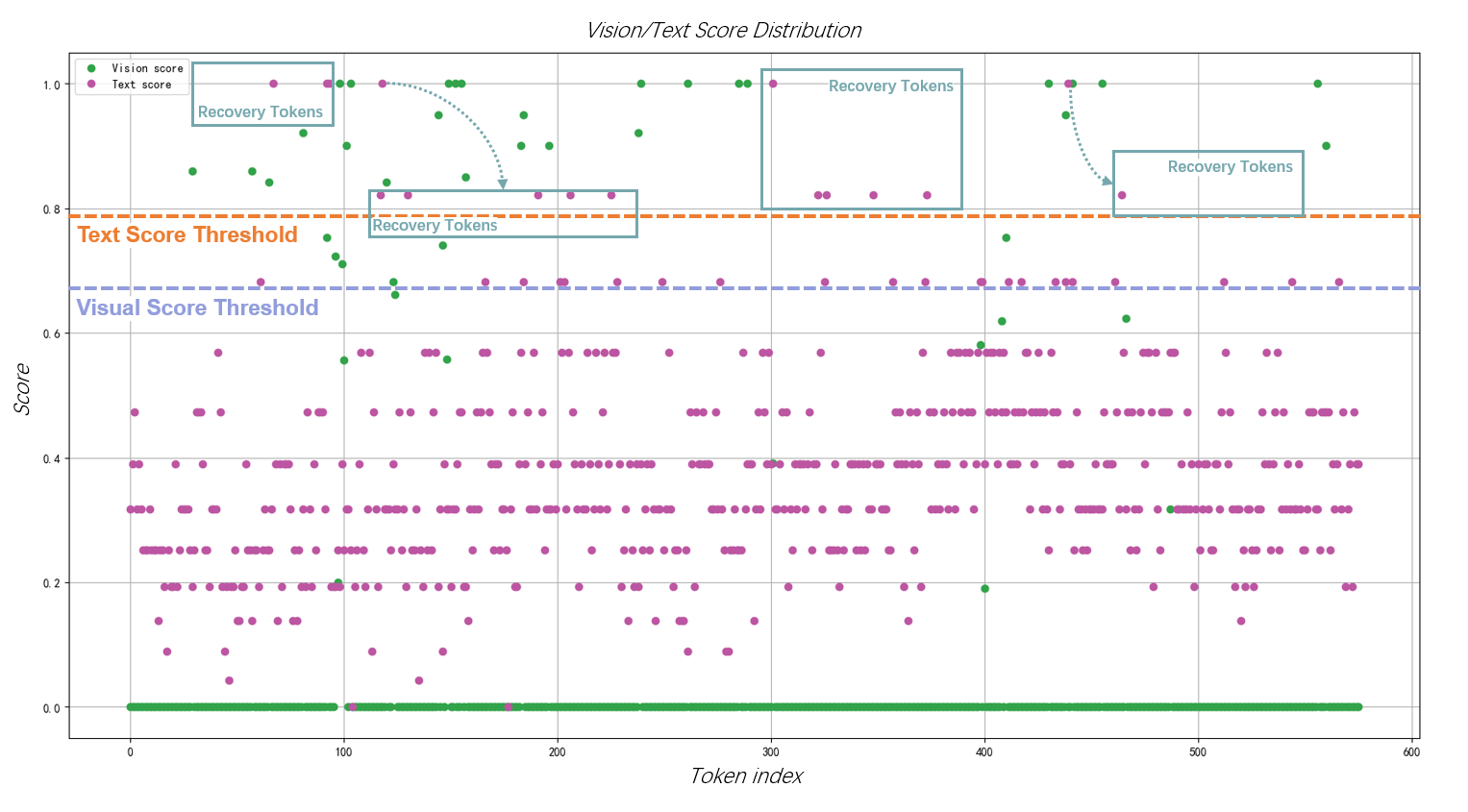}
\caption{An example of visual and text scores distribution, where the box enclosed represents the area needed to be recovered with text information.} 
\label{fig: sample}
\end{figure*}

Finally, the assembled visual tokens are projected into the semantic domain of the LLM via a dedicated projection layer, subsequently being fed into the LLM in conjunction with the input text.

\subsection{Multimodal Score}
\textbf{Visual Score\quad} In the ViT model, in addition to representing visual labels for image patches, a separate class label is also introduced. Class labeling is obtained by aggregating information from all visual labels using a global attention mechanism, and it captures the global representation of the image. Therefore, the dot product between class tags and visual tags can represent the importance of each visual token relative to the global context of the image \cite{liang2022not,chen2023diffrate}. Specifically, we define $W_{cls}$ as the representation vector matrix for the class token and $W_{token}$ as the representation vector matrix for the other visual tokens. Then, the visual score can be defined as:
\begin{equation}
    Score_{v} = Softmax(\frac{W_{cls} \cdot W_{token}^T}{\sqrt{d_{W_{cls}}}}).
\end{equation}

Where $d_{W_{cls}}$ represents the magnitude of $W_{cls}$ (i.e., the length of the vector).

Visual tokens with higher $Score_{v}$ represent a stronger correlation with the global features, indicating a higher similarity to the overall image. Therefore, $Score_{v}$ can be used to measure the importance of each visual token in the visual modality.

\textbf{Text Score\quad} Similar to the visual modality, in a transformer model for the text modality, the model also learns features for each text token and integrates them into a global text embedding using the global attention mechanism, capturing the global context of the text. The dot product between the text embedding and visual tokens can represent the importance of each visual token relative to the global context of the text. However, due to the modality gap between text and visual modalities, directly using the dot product between each visual token and the text embedding as a measure of text-visual similarity is not appropriate. Therefore, we calculate the similarity between the projected visual tokens and the text embedding. We define $W_{text}$ as the representation vector matrix for the text embedding, $W_{token}$ as the representation vector matrix for the other visual tokens, $MLP$ represents a projection layer that aligns two modalities, Then, the text score can be defined as:

\begin{equation}
    Score_{t} = Softmax(\frac{W_{text} \cdot MLP(W_{token})^T}{\sqrt{d_{W_{text}}}}).
\end{equation}

Where $d_{W_{text}}$ represents the magnitude of $W_{text}$ (i.e., the length of the vector).

First, we use $Score_v$ to select the most important tokens in the visual modality, ensuring the effectiveness of visual information. Then, we use $Score_t$ to recover the tokens that were filtered out in the first step. This step aims to retrieve the tokens that contain helpful information for answering the question, which may have been lost during the initial filtering. As shown in Figure \ref{fig: sample}, the tokens enclosed in the boxes represent the visual score as zero but have relatively high textual scores. These tokens may have a strong relevance to the textual question and need to be recovered with the text score. Finally, we apply the KNN to merge the remaining tokens, ensuring that background and other contextual information are not heavily lost. This is because in some cases, background information also contains visually relevant information that can be useful for answering the question. Through these two rounds of token recovery, visual, semantic, and background information are preserved more comprehensively.

\subsection{Dynamic Scale Filter}
After obtaining the similarity scores between the two modalities, we conducted visual analysis as shown in Figure \ref{fig: sample}. We normalized both visual and textual scores to the range of $[0, 1]$. Both visual and textual scores exhibit prominent data points that contain more valuable information for the respective modality. By preserving the tokens corresponding to these important data points and merging the remaining tokens, we can ensure the retention of as much complete and valuable information as possible while reducing the number of tokens. From a data distribution perspective, these data points containing more valuable information can be considered outliers. Therefore, we transform the task of selecting informative tokens into detecting outliers.

Due to the variation in content for each instance, it is not reasonable to use a fixed threshold for outlier selection. Some instances may contain a small number of important tokens, far below the fixed threshold, and using a fixed threshold for selection would result in additional computational overhead. On the other hand, some instances may contain a large number of important tokens, far exceeding the fixed threshold, and using a fixed threshold would lead to a loss of significant information. Therefore, it is necessary for the dynamic scale filtering method to dynamically adjust based on the data distribution of each instance.

To dynamically detect the outliers that contain more valuable information, we utilized the Local Outlier Factor (LOF) \cite{breunig2000lof}. This is a classical density-based outlier detection algorithm that computes the ratio of the density of each data point to the density of its surrounding neighborhood points. 

Specifically, define the K-nearest neighbor distance, which represents the distance between the $k$-th point and the current data point \(P\), denoted as \(d_{k}(P) = d(P, O)\). At this, with \(P\) as the center and \(d_{k}(P)\) as the radius, we define a circle. The range encompassed by this circle is called the K-distance neighborhood, denoted as \(N_{k}(P) = \{d(P, O') \leq d_{k}(P)\}\). Then, the reachable distance $Reachdist_{k}(O, P)=\max \{d_{k}(O),d(O, P)\}$ measures the density of the surrounding points of $P$ with respect to the neighboring point $O$. According to the above, local reachability density can be defined as:
\begin{equation}
\operatorname{LRD}_k(P)=\frac{1}{\frac{\sum_{O_\ni N_k(P)} \text {Reachdist}(P, O)}{\left|N_k(P)\right|}}
\label{reachability}
\end{equation}
Based on Equation \ref{reachability}, the local outlier factor (LOF) can be defined as:
\begin{equation}
\begin{aligned}
\operatorname{LOF}_k(P) &=\frac{\sum_{O \ni N_k(P)} \frac{\operatorname{LRD}(O)}{\operatorname{LRD}(P)}}{\left|N_k(P)\right|}\\
&=\frac{\sum_{O \ni N_k(P)} \operatorname{LRD}(O)}{\left|N_k(P)\right|} / \operatorname{LRD} d(P)
\label{lof}
\end{aligned}
\end{equation}
For equation \ref{lof}, the following conclusions can be drawn:
\begin{equation}
\begin{cases}
    \operatorname{LOF}_k(P) \leq 1, \text{ Other (Normal).} \\
    \operatorname{LOF}_k(P) > 1, \text{ Important (Outlier).}
\end{cases}
\end{equation}
By using this algorithm, we can identify the proportion of tokens with exceptional scores (i.e., tokens containing more valuable information) among all the tokens, enabling us to perform the selection process.

\begin{figure*}[!htb]
\centering
\includegraphics[width=0.8\linewidth]{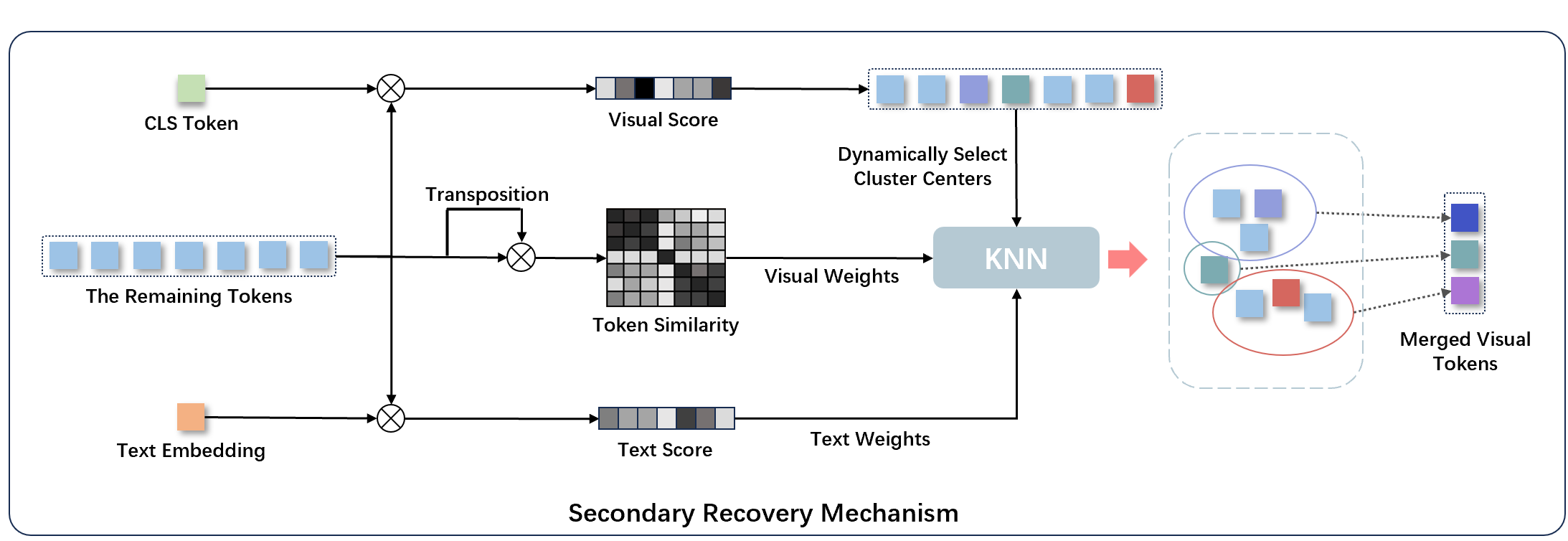}
\caption{Framework of secondary recovery mechanism based on visual and text
information.} 
\label{fig: second}
\end{figure*}  

\subsection{Token Secondary Recovery}
After extracting text information for recovery, most of the remaining tokens are associated with the background of the image. For some instances of the problem, the background information is not useful. However, in other instances, the background information may contain valuable insights for problem-solving. Simply discarding these tokens would result in the loss of valuable information. To address this, we employ the KNN algorithm to cluster the remaining tokens, thereby performing a second round of token recovery that preserves the background information.

As shown in Figure \ref{fig: second}, for the remaining tokens after the first round of recovery, we still use the dot product with category tokens as the visual score. Then, we apply the same approach as the previous section but with different parameters to filter outlier tokens. These outliers are considered the initial cluster centers because they still contain relatively high useful information within the remaining tokens. The dot product between each pair of tokens is used as the distance metric during clustering. Finally, the tokens within each cluster are merged.

\section{Experiments}
\subsection{Datasets}
We evaluated our method on the following publicly and widely available multimodal datasets:  \textbf{ScienceQA} \cite{lu2022learn}, \textbf{TextVQA} \cite{singh2019towards}, \textbf{MME} \cite{fu2023mme}, \textbf{VQAv2} \cite{goyal2017making}, \textbf{POPE} \cite{li2023evaluating} and \textbf{MMBench} \cite{liu2023mmbench}. 

\subsection{Implementation Details}
All experiments were conducted in the PyTorch framework on four NVIDIA 4090 24G GPUs. We utilized the \textbf{LLaVA1.5-7B} with LoRA fine-tuning as our baseline. The visual and text encoder is CLIP with ViT as the backbone, where the input image size is 336x336 and the patch size is 14x14. It's worth noting that we followed the CLIP pretraining approach for handling images of different resolutions, directly resizing them to 336x336. The parameter $k$ in LOF is set to $20$ in ScienceQA, TextVQA, and MMBench. In the MME, VQAv2, and POPE $k$ was set to $30$, $90$, and $30$ respectively. In the fine-tuning experiment, we used LoRA\cite{hu2022lora} to fine-tune LLaVA1.5-7B and train one epoch with the same data.

\subsection{Ablation Study}
\subsubsection{Efficiency Analysis for Dynamic Scale Filter}
To verify the effectiveness of the dynamic threshold, we conducted a control experiment using the top $k\%$ tokens ranked by visual scores. The results are shown in Table \ref{tab: dynamic threshold}. The ratio between the number of tokens used and the total number of tokens is indicated in parentheses. In this experiment, the dynamic threshold method has shown better performance by dynamically selecting thresholds based on different instances. Due to variations in the distribution of visual information across different instances, using a simple fixed threshold for filtering would result in different losses for different instances. On the one hand, some instances can be effectively answered using a token count lower than the fixed threshold, and having extra tokens in such cases would lead to unnecessary computational overhead. On the other hand, for certain instances, the model requires more tokens than the fixed threshold to answer questions accurately, resulting in information loss with a fixed threshold. By adopting the dynamic threshold method, both scenarios can be avoided, achieving a balance between the number of tokens and performance.   
\begin{table}[!htb]
\centering
\caption{The effectiveness of the dynamic scale filter.}
\label{tab: dynamic threshold}
\begin{tabular}{@{}lllllll@{}}
\toprule
Method & \multicolumn{2}{l}{ScienceQA}     & \multicolumn{2}{l}{TextVQA}     & \multicolumn{2}{l}{MME} \\ \midrule
    Fixed  & \multicolumn{2}{l}{68.47 (5.7\%)} & \multicolumn{2}{l}{54.20 (6.5\%)}      & \multicolumn{2}{l}{1139.2 (5.4\%)}    \\ \midrule
    Dynamic    & \multicolumn{2}{l}{\textbf{68.57 (5.7\%)}} & \multicolumn{2}{l}{\textbf{54.60 (6.5\%)}}  & \multicolumn{2}{l}{\textbf{1146.9 (5.4\%)}} \\ \bottomrule
\end{tabular}
\end{table}

\subsubsection{Efficiency Analysis for Text Information}
To ensure the effectiveness of text information, we conducted three comparative experiments, two of which were directly screened based on visual scores, and the remaining using visual score screening and recovery using text information. As shown in Table \ref{tab: text information}. When using a similar proportion of token counts, the method of utilizing text information for recovery demonstrates better performance. Under the setting of utilizing text information for recovery, using $8.7\%$ of the token count achieves better performance on all tasks compared to directly using visual scores for filtering with $10.2\%$ of the tokens. This is because text information allows the model to better focus on areas related to the question. There is also an interesting observation that for the ScienceQA task, the relationship between token count and performance is not intuitive. This is due to the presence of redundancy and interference in tokens in this task, resulting in inconsistent trends in token count and performance improvement.
\begin{table}[!htb]
\small
\centering
\fontsize{9pt}{10pt}\selectfont
\caption{The effectiveness of the text information.}
\label{tab: text information}
\begin{tabular}{@{}lllllll@{}}
\toprule
Method        & \multicolumn{2}{l}{ScienceQA}      & \multicolumn{2}{l}{TextVQA}        & \multicolumn{2}{l}{MME} \\ \midrule
Top 10.2\% & \multicolumn{2}{l}{68.52 (10.2\%)} & \multicolumn{2}{l}{55.22 (10.2\%)} & \multicolumn{2}{l}{1194.9 (10.2\%)}    \\
Top 8.7\%  & \multicolumn{2}{l}{68.72 (8.7\%)}  & \multicolumn{2}{l}{54.94 (8.7\%)}    & \multicolumn{2}{l}{1190.3 (8.7\%)}    \\ \midrule
Vision + Text & \multicolumn{2}{l}{\textbf{68.91 (8.7\%)}} & \multicolumn{2}{l}{\textbf{55.33 (8.7\%)}}             & \multicolumn{2}{l}{\textbf{1196.9 (8.8\%)}} \\ \bottomrule
\end{tabular}
\end{table}

\begin{table}[!htb]
\centering
\caption{The effectiveness of the second recovery mechanism.}
\label{tab: second recovery mechanism}
\begin{tabular}{@{}lllllll@{}}
\toprule
Method      & \multicolumn{2}{l}{ScienceQA}              & \multicolumn{2}{l}{TextVQA}                      & \multicolumn{2}{l}{MME}       \\ \midrule
Single        & \multicolumn{2}{l}{68.91 (8.7\%)}          & \multicolumn{2}{l}{55.33 (8.7\%)}                   & \multicolumn{2}{l}{1196.9 (8.8\%)}          \\ \midrule
Second  & \multicolumn{2}{l}{\textbf{69.01 (9.7\%)}} & \multicolumn{2}{l}{\textbf{55.51 (9.9\%)}}  & \multicolumn{2}{l}{\textbf{1284.9 (9.2\%)}} \\ \bottomrule
\end{tabular}
\end{table}

\subsubsection{Efficiency Analysis for Secondary Recovery Mechanism}
To validate the effectiveness of the second recovery, we conducted experiments with the same settings and parameters. As shown in Table \ref{tab: second recovery mechanism}, the results demonstrate that further performance improvement can be achieved by merging the remaining tokens, as it helps capture beneficial information from the background for certain instances.

\begin{table}[!htb]
\centering
\caption{The effectiveness of the compression ratio}
\label{tab:compression ratio}
\begin{tabular}{@{}lllllll@{}}
\toprule
Ratio      & \multicolumn{2}{l}{ScienceQA}              & \multicolumn{2}{l}{TextVQA}                      & \multicolumn{2}{l}{MME}       \\ \midrule
5\%        & \multicolumn{2}{l}{67.77 (4.5\%)}          & \multicolumn{2}{l}{51.04 (4.6\%)}                   & \multicolumn{2}{l}{1187.9 (4.8\%)}     \\ \midrule  
10\%  & \multicolumn{2}{l}{\textbf{69.01 (9.7\%)}} & \multicolumn{2}{l}{55.51 (9.9\%)}  & \multicolumn{2}{l}{1284.9 (9.2\%)} \\ \midrule 
50\%  & \multicolumn{2}{l}{68.52 (49.5\%)} & \multicolumn{2}{l}{\textbf{56.14 (51.1\%)}}  & \multicolumn{2}{l}{\textbf{1301.8 (50.9\%)}} \\
\bottomrule
\end{tabular}
\end{table}

\subsubsection{Efficiency Analysis for Compression Ratio}
To verify the impact of different compression ratios on performance, As shown in \ref{tab:compression ratio}, We conducted comparisons at compression rates of approximately 5\%, 10\%, and 50\%, The results indicate that maintaining a compression rate of around 10\% can ensure both efficiency and performance simultaneously.

\begin{table*}[!htb]
\centering
\caption{Performance comparison with other multimodal models and pruning methods.}
\label{tab: result}
\begin{tabular}{lllllllll}
\hline
Method         & ScienceQA  & TextVQA  & MME        & VQAv2      & POPE    & MMBench  \\ \hline
BLIP-2         & 61.00      & 42.50    & 1293.80    & 41.00      & 85.30   & -       \\
InstrucBILP    & 60.50      & 50.10    & -          & -          & -       & 36.00   \\
InstrucBILP    & 63.10      & 50.70    & 1212.80    & -          & 78.90   & -       \\
Shikra         & -          & -        & -          & 77.40      & -       & 58.80    \\
IDEFICS-9B     & -          & 25.90    & -          & 50.90      & -       & 48.20     \\
IDEFICS-80B    & -          & 30.90    & -          & 60.00      & -       & 54.50     \\
Qwen-VL        & 67.10      & \textbf{63.80}    & -     & 78.80      & -       & 38.20    \\
LLaVA-1.5      & \textbf{68.40}      & 58.20    & \textbf{1476.90}    & \textbf{79.10}      & \textbf{86.40}   & \textbf{66.10}    \\ \hline
\textcolor{gray}{Fine-tuning Method} \\
LLaVA-PruMerge & 68.50     & 56.00    & 1350.30    & 72.00      & 76.30   & 60.90     \\
LLaVA-PruMerge+ & 68.30      & \textbf{57.10}    & 1462.40    & 76.80      & \textbf{84.00}   & \textbf{64.90}     \\
CrossGET & 66.70      & 54.90    & \textbf{1510.20}    & \textbf{77.30}      & 83.90   & 64.70    \\
Chat-UniVi &59.96   &- &- &- &73.10 &-
\\ 
Ours & \textbf{68.72} &56.16 &1323.54 &71.18 &79.50 &59.20 \\
\hline
\textcolor{gray}{Training-Free Method} \\
ToMe           & 50.00      & 45.30    & 1138.00    & 57.10      & 52.50   & 43.70 \\
LLaVA-PruMerge & 68.52      & 53.51    & 1191.50    & 65.90      & 70.70   & 56.78     \\
Ours           & \textbf{69.01}      & \textbf{55.51}    & \textbf{1284.90}    & \textbf{70.41}      & \textbf{72.00}   & \textbf{57.90}   \\ \hline
\end{tabular}
\end{table*}
\begin{table*}[!htb]
\centering
\caption{Comparison of computational costs on NVIDIA A100 GPU.}
\label{tab: costs}
\begin{tabular}{llccccc}
\hline
Method &
  \multicolumn{1}{c}{\begin{tabular}[c]{@{}c@{}}LLM \\ Backbone\end{tabular}} &
  Quantization &
  \begin{tabular}[c]{@{}c@{}}FLOPs\\ (T)\end{tabular} &
  \begin{tabular}[c]{@{}c@{}}Prefill\\ Time (ms)\end{tabular} &
  \begin{tabular}[c]{@{}c@{}}Total\\ Memory (G)\end{tabular} &
  \begin{tabular}[c]{@{}c@{}}Storing\\ Activation (G)\end{tabular} \\ \hline
LLaVA1.5 & \multicolumn{1}{c}{Vicuna-7B} & FP16 & 8.5          & 30.3         & 22.2          & 4.1           \\
Ours     & \multicolumn{1}{c}{Vicuna-7B} & FP16 & \textbf{1.5} & \textbf{9.2} & \textbf{14.4} & \textbf{0.49} \\ \hline
LLaVA1.5 & Vicuna-7B                     & INT8 & 4.3          & 15.2         & 11.1          & 2.0           \\
Ours     & Vicuna-7B                     & INT8 & \textbf{0.8} & \textbf{4.6} & \textbf{7.2}  & \textbf{0.24} \\ \hline
LLaVA1.5 & Vicuna-7B                     & INT4 & 2.1          & 14.2         & 5.56          & 1.0           \\
Ours     & Vicuna-7B                     & INT4 & \textbf{0.4} & \textbf{2.6} & \textbf{3.6}  & \textbf{0.12} \\ \hline
\end{tabular}%
\end{table*}

\subsection{Main Results}
To further validate the effectiveness of our method, we implemented our approach based on LLaVA1.5 and conducted comparative experiments with other MM-LLMs and existing MM-LLM token pruning methods. The results are presented in Table \ref{tab: result}.

According to the results, our method achieves usable performance even when using only $10\%$ of the average number of tokens, especially in the ScienceQA and TextVQA. This is because these two tasks require the model to focus more on areas with high relevance to the problem text, which is consistent with the expectation of our method. Using text information to retrieve visual tokens with high similarity to the problem, helps the model maintain good or even better performance while reducing computational complexity.
In other tasks, our method shows a decrease in performance, because these tasks require more raw visual features. As our method is dynamic, we can control the number of raw visual tokens by adjusting the $k$ in LOF to ensure competitive performance.
Compared to other training-free token pruning methods for MM-LLM, our method demonstrates strong competitiveness. Despite having a similar order of magnitude in terms of token count, our proposed method outperforms others in performance. Compared with fine-tuning methods, our method is still competitive. CrossGET is also an acceleration method for text-visual modality interaction, but unlike it, our method preserves the original visual tokens highly associated with the text during pruning. Compared with merged tokens, the model has a stronger understanding of the original tokens, so our method performs better on ScienceQA and TextVQA. 

At the same time, we conducted fine-tuning experiments under the setting of ensuring that the average number of tokens used is 10\% of the original number of tokens, The experimental results indicate that our proposed method achieved further performance improvements following fine-tuning. Compared to LLaVA PruMerge, we achieved better performance on most datasets when using a similar magnitude of tokens. At the same time, our method does not require specific hyperparameter tuning for the datasets and demonstrates better generalization capabilities. Compared to CrossGET, we narrowed the performance gap following fine-tuning. When compared to the original LLaVA1.5-7B, our method reduces computational overhead by 82\%, while CrossGET achieves a reduction of 31\%. Furthermore, compared to LLaVA-PruMerge+, the number of tokens we utilized is half of theirs, yet we still achieved competitive performance across multiple datasets. Chat-UniVi\cite{jin2024chat} is a unified vision-language model capable of comprehending and engaging in conversations involving images and videos through a unified visual representation, which uses the strategy in ToMe\cite{bolya2023token}, also try to compress visual token in MM-LMM. Compared to it and ToMe, our method exhibits better performance. This stems from the fact that our method is tailored for large multimodal models, simultaneously considering crucial information from both the visual and text modalities.

We analyze the computational cost of our method using an open-source tool \cite{yuan2024llm} on the NVIDIA A100 GPU. Assuming the text input length of $60$. As shown in Table \ref{tab: costs}, compared with the base model, our method significantly reduces computational and memory consumption while ensuring good usability performance.

\subsection{Visualization}
\begin{figure}[!htb]
\centering
\includegraphics[width=1.\linewidth]{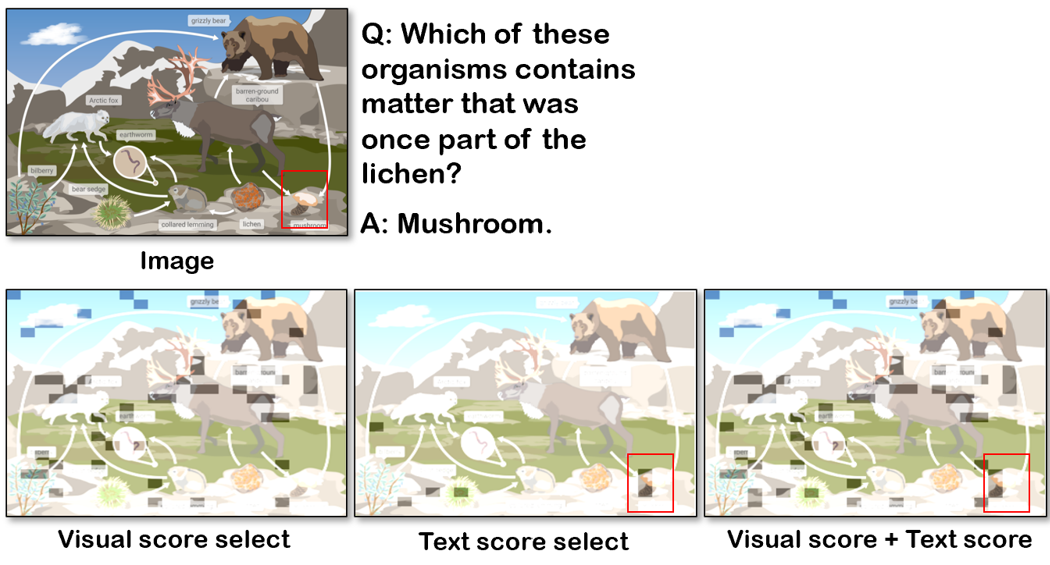}
\caption{Visualization results of token select/recovery with visual/text scores. The red box area represents the tokens corresponding to the answer.} 
\label{fig: v1}
\end{figure}  

As shown in Figure \ref{fig: v1}, The tokens used for visual score screening are disorganized and do not contain the image regions corresponding to the final answer. The tokens collected for text information recovery are orderly, concentrated in regions related to the question, and include the regions contained in the answer. This indicates that our proposed method can recover lost important information through textual information. 
More visualization results can be found in the Appendix-1.

\section{Conclusion}

In this paper, we propose a multimodal visual token recovery mechanism guided by text information, which retains as much information as possible by reclaiming important visual tokens through textual information. Additionally, it consolidates background information using KNN to achieve efficient inference in MM-LLM.
Our approach achieves competitive performance on multiple tasks and provides valuable insights and methods for efficient MM-LLM.

\section{Limitation}
There is still room for improvement in the performance of our method, we will enhance its performance further and adapt this in multiple rounds of VQA tasks in the future.

\section{Acknowledgements}
This work was supported by the Strategic Priority Research Program of the Chinese Academy of Sciences (No.XDA0450202), Beijing Municipal Science and Technology Project (No.Z231100010323005), CAS Project for Young Scientists in Basic Research (YSBR-083), and 2035 Innovation Mission Project (No.E4J10102).

\bibliography{aaai25}

\begin{thebibliography}{37}
\providecommand{\natexlab}[1]{#1}

\bibitem[{Bolya et~al.(2023)Bolya, Fu, Dai, Zhang, Feichtenhofer, and Hoffman}]{bolya2023token}
Bolya, D.; Fu, C.-Y.; Dai, X.; Zhang, P.; Feichtenhofer, C.; and Hoffman, J. 2023.
\newblock Token Merging: Your ViT But Faster.
\newblock In \emph{The Eleventh International Conference on Learning Representations}.

\bibitem[{Breunig et~al.(2000)Breunig, Kriegel, Ng, and Sander}]{breunig2000lof}
Breunig, M.~M.; Kriegel, H.-P.; Ng, R.~T.; and Sander, J. 2000.
\newblock LOF: identifying density-based local outliers.
\newblock In \emph{Proceedings of the 2000 ACM SIGMOD international conference on Management of data}, 93--104.

\bibitem[{Brock et~al.(2021)Brock, De, Smith, and Simonyan}]{brock2021high}
Brock, A.; De, S.; Smith, S.~L.; and Simonyan, K. 2021.
\newblock High-performance large-scale image recognition without normalization.
\newblock In \emph{International Conference on Machine Learning}, 1059--1071. PMLR.

\bibitem[{Brown et~al.(2020)Brown, Mann, Ryder, Subbiah, Kaplan, Dhariwal, Neelakantan, Shyam, Sastry, Askell et~al.}]{brown2020language}
Brown, T.; Mann, B.; Ryder, N.; Subbiah, M.; Kaplan, J.~D.; Dhariwal, P.; Neelakantan, A.; Shyam, P.; Sastry, G.; Askell, A.; et~al. 2020.
\newblock Language models are few-shot learners.
\newblock \emph{Advances in neural information processing systems}, 33: 1877--1901.

\bibitem[{Cao et~al.(2024)Cao, Ye, Li, Yu, Tang, Lu, and Chen}]{cao2024madtp}
Cao, J.; Ye, P.; Li, S.; Yu, C.; Tang, Y.; Lu, J.; and Chen, T. 2024.
\newblock MADTP: Multimodal Alignment-Guided Dynamic Token Pruning for Accelerating Vision-Language Transformer.
\newblock In \emph{Proceedings of the IEEE/CVF conference on computer vision and pattern recognition}.

\bibitem[{Chen et~al.(2023)Chen, Shao, Xu, Lin, Zhang, Chao, Ji, Qiao, and Luo}]{chen2023diffrate}
Chen, M.; Shao, W.; Xu, P.; Lin, M.; Zhang, K.; Chao, F.; Ji, R.; Qiao, Y.; and Luo, P. 2023.
\newblock Diffrate: Differentiable compression rate for efficient vision transformers.
\newblock In \emph{Proceedings of the IEEE/CVF International Conference on Computer Vision}, 17164--17174.

\bibitem[{Chu et~al.(2023)Chu, Qiao, Lin, Xu, Yang, Hu, Wei, Zhang, Zhang, Wei et~al.}]{chu2023mobilevlm}
Chu, X.; Qiao, L.; Lin, X.; Xu, S.; Yang, Y.; Hu, Y.; Wei, F.; Zhang, X.; Zhang, B.; Wei, X.; et~al. 2023.
\newblock Mobilevlm: A fast, reproducible and strong vision language assistant for mobile devices.
\newblock \emph{arXiv preprint arXiv:2312.16886}.

\bibitem[{Dosovitskiy et~al.(2021)Dosovitskiy, Beyer, Kolesnikov, Weissenborn, Zhai, Unterthiner, Dehghani, Minderer, Heigold, Gelly, Uszkoreit, and Houlsby}]{dosovitskiy2020image}
Dosovitskiy, A.; Beyer, L.; Kolesnikov, A.; Weissenborn, D.; Zhai, X.; Unterthiner, T.; Dehghani, M.; Minderer, M.; Heigold, G.; Gelly, S.; Uszkoreit, J.; and Houlsby, N. 2021.
\newblock An Image is Worth 16x16 Words: Transformers for Image Recognition at Scale.
\newblock \emph{ICLR}.

\bibitem[{Fu et~al.(2023)Fu, Chen, Shen, Qin, Zhang, Lin, Yang, Zheng, Li, Sun, Wu, and Ji}]{fu2023mme}
Fu, C.; Chen, P.; Shen, Y.; Qin, Y.; Zhang, M.; Lin, X.; Yang, J.; Zheng, X.; Li, K.; Sun, X.; Wu, Y.; and Ji, R. 2023.
\newblock MME: A Comprehensive Evaluation Benchmark for Multimodal Large Language Models.
\newblock \emph{arXiv preprint arXiv:2306.13394}.

\bibitem[{Ganz et~al.(2024)Ganz, Kittenplon, Aberdam, Avraham, Nuriel, Mazor, and Litman}]{ganz2024question}
Ganz, R.; Kittenplon, Y.; Aberdam, A.; Avraham, E.~B.; Nuriel, O.; Mazor, S.; and Litman, R. 2024.
\newblock Question Aware Vision Transformer for Multimodal Reasoning.
\newblock \emph{arXiv preprint arXiv:2402.05472}.

\bibitem[{Goyal et~al.(2017)Goyal, Khot, Summers-Stay, Batra, and Parikh}]{goyal2017making}
Goyal, Y.; Khot, T.; Summers-Stay, D.; Batra, D.; and Parikh, D. 2017.
\newblock Making the v in vqa matter: Elevating the role of image understanding in visual question answering.
\newblock In \emph{Proceedings of the IEEE conference on computer vision and pattern recognition}, 6904--6913.

\bibitem[{Hu et~al.(2022)Hu, yelong shen, Wallis, Allen-Zhu, Li, Wang, Wang, and Chen}]{hu2022lora}
Hu, E.~J.; yelong shen; Wallis, P.; Allen-Zhu, Z.; Li, Y.; Wang, S.; Wang, L.; and Chen, W. 2022.
\newblock Lo{RA}: Low-Rank Adaptation of Large Language Models.
\newblock In \emph{International Conference on Learning Representations}.

\bibitem[{Jiang et~al.(2023)Jiang, Sablayrolles, Mensch, Bamford, Chaplot, Casas, Bressand, Lengyel, Lample, Saulnier et~al.}]{jiang2023mistral}
Jiang, A.~Q.; Sablayrolles, A.; Mensch, A.; Bamford, C.; Chaplot, D.~S.; Casas, D. d.~l.; Bressand, F.; Lengyel, G.; Lample, G.; Saulnier, L.; et~al. 2023.
\newblock Mistral 7B.
\newblock \emph{arXiv preprint arXiv:2310.06825}.

\bibitem[{Jin et~al.(2024)Jin, Takanobu, Zhang, Cao, and Yuan}]{jin2024chat}
Jin, P.; Takanobu, R.; Zhang, W.; Cao, X.; and Yuan, L. 2024.
\newblock Chat-univi: Unified visual representation empowers large language models with image and video understanding.
\newblock In \emph{Proceedings of the IEEE/CVF Conference on Computer Vision and Pattern Recognition}, 13700--13710.

\bibitem[{Kong et~al.(2022)Kong, Dong, Ma, Meng, Niu, Sun, Shen, Yuan, Ren, Tang et~al.}]{kong2022spvit}
Kong, Z.; Dong, P.; Ma, X.; Meng, X.; Niu, W.; Sun, M.; Shen, X.; Yuan, G.; Ren, B.; Tang, H.; et~al. 2022.
\newblock Spvit: Enabling faster vision transformers via latency-aware soft token pruning.
\newblock In \emph{European conference on computer vision}, 620--640. Springer.

\bibitem[{Lee, Choi, and Kim(2024)}]{lee2024multi}
Lee, S.; Choi, J.; and Kim, H.~J. 2024.
\newblock Multi-criteria Token Fusion with One-step-ahead Attention for Efficient Vision Transformers.
\newblock In \emph{Proceedings of the IEEE/CVF Conference on Computer Vision and Pattern Recognition}.

\bibitem[{Li et~al.(2023)Li, Du, Zhou, Wang, Zhao, and Wen}]{li2023evaluating}
Li, Y.; Du, Y.; Zhou, K.; Wang, J.; Zhao, W.~X.; and Wen, J.-R. 2023.
\newblock Evaluating Object Hallucination in Large Vision-Language Models.
\newblock In \emph{Proceedings of the 2023 Conference on Empirical Methods in Natural Language Processing}, 292--305.

\bibitem[{Liang et~al.(2022)Liang, Ge, Tong, Song, Xie et~al.}]{liang2022not}
Liang, Y.; Ge, C.; Tong, Z.; Song, Y.; Xie, P.; et~al. 2022.
\newblock Not all patches are what you need: Expediting vision transformers via token reorganizations.
\newblock In \emph{International Conference on Learning Representations (ICLR)}.

\bibitem[{Liu et~al.(2024{\natexlab{a}})Liu, Yin, Cao, Jiang, Li, Liu, Jiang, Sun, and Xu}]{liu2024hrvda}
Liu, C.; Yin, K.; Cao, H.; Jiang, X.; Li, X.; Liu, Y.; Jiang, D.; Sun, X.; and Xu, L. 2024{\natexlab{a}}.
\newblock HRVDA: High-Resolution Visual Document Assistant.
\newblock In \emph{Proceedings of the IEEE/CVF conference on computer vision and pattern recognition}.

\bibitem[{Liu et~al.(2023{\natexlab{a}})Liu, Li, Li, and Lee}]{liu2023improved}
Liu, H.; Li, C.; Li, Y.; and Lee, Y.~J. 2023{\natexlab{a}}.
\newblock Improved Baselines with Visual Instruction Tuning.
\newblock In \emph{NeurIPS 2023 Workshop on Instruction Tuning and Instruction Following}.

\bibitem[{Liu et~al.(2024{\natexlab{b}})Liu, Li, Li, Li, Zhang, Shen, and Lee}]{liu2024llava}
Liu, H.; Li, C.; Li, Y.; Li, B.; Zhang, Y.; Shen, S.; and Lee, Y.~J. 2024{\natexlab{b}}.
\newblock LLaVA-NeXT: Improved reasoning, OCR, and world knowledge.

\bibitem[{Liu et~al.(2023{\natexlab{b}})Liu, Duan, Zhang, Li, Zhang, Zhao, Yuan, Wang, He, Liu et~al.}]{liu2023mmbench}
Liu, Y.; Duan, H.; Zhang, Y.; Li, B.; Zhang, S.; Zhao, W.; Yuan, Y.; Wang, J.; He, C.; Liu, Z.; et~al. 2023{\natexlab{b}}.
\newblock Mmbench: Is your multi-modal model an all-around player?
\newblock \emph{arXiv preprint arXiv:2307.06281}.

\bibitem[{Long et~al.(2023)Long, Zhao, Pi, Wang, and Wang}]{long2023beyond}
Long, S.; Zhao, Z.; Pi, J.; Wang, S.; and Wang, J. 2023.
\newblock Beyond attentive tokens: Incorporating token importance and diversity for efficient vision transformers.
\newblock In \emph{Proceedings of the IEEE/CVF Conference on Computer Vision and Pattern Recognition}, 10334--10343.

\bibitem[{Lu et~al.(2022)Lu, Mishra, Xia, Qiu, Chang, Zhu, Tafjord, Clark, and Kalyan}]{lu2022learn}
Lu, P.; Mishra, S.; Xia, T.; Qiu, L.; Chang, K.-W.; Zhu, S.-C.; Tafjord, O.; Clark, P.; and Kalyan, A. 2022.
\newblock Learn to explain: Multimodal reasoning via thought chains for science question answering.
\newblock \emph{Advances in Neural Information Processing Systems}, 35: 2507--2521.

\bibitem[{OpenAI(2023)}]{openai2023gpt}
OpenAI. 2023.
\newblock GPT-4V (ision) System Card.
\newblock \emph{Citekey: gptvision}.

\bibitem[{Ouyang et~al.(2022)Ouyang, Wu, Jiang, Almeida, Wainwright, Mishkin, Zhang, Agarwal, Slama, Ray et~al.}]{ouyang2022training}
Ouyang, L.; Wu, J.; Jiang, X.; Almeida, D.; Wainwright, C.; Mishkin, P.; Zhang, C.; Agarwal, S.; Slama, K.; Ray, A.; et~al. 2022.
\newblock Training language models to follow instructions with human feedback.
\newblock \emph{Advances in neural information processing systems}, 35: 27730--27744.

\bibitem[{Radford et~al.(2021)Radford, Kim, Hallacy, Ramesh, Goh, Agarwal, Sastry, Askell, Mishkin, Clark et~al.}]{radford2021learning}
Radford, A.; Kim, J.~W.; Hallacy, C.; Ramesh, A.; Goh, G.; Agarwal, S.; Sastry, G.; Askell, A.; Mishkin, P.; Clark, J.; et~al. 2021.
\newblock Learning transferable visual models from natural language supervision.
\newblock In \emph{International conference on machine learning}, 8748--8763. PMLR.

\bibitem[{Rao et~al.(2021)Rao, Zhao, Liu, Lu, Zhou, and Hsieh}]{rao2021dynamicvit}
Rao, Y.; Zhao, W.; Liu, B.; Lu, J.; Zhou, J.; and Hsieh, C.-J. 2021.
\newblock Dynamicvit: Efficient vision transformers with dynamic token sparsification.
\newblock \emph{Advances in neural information processing systems}, 34: 13937--13949.

\bibitem[{Shang et~al.(2024)Shang, Cai, Xu, Lee, and Yan}]{shang2024llava}
Shang, Y.; Cai, M.; Xu, B.; Lee, Y.~J.; and Yan, Y. 2024.
\newblock LLaVA-PruMerge: Adaptive Token Reduction for Efficient Large Multimodal Models.
\newblock \emph{arXiv preprint arXiv:2403.15388}.

\bibitem[{Shi et~al.(2024)Shi, Tao, Rao, Yang, Yuan, and Wang}]{shi2023crossget}
Shi, D.; Tao, C.; Rao, A.; Yang, Z.; Yuan, C.; and Wang, J. 2024.
\newblock Crossget: Cross-guided ensemble of tokens for accelerating vision-language transformers.
\newblock In \emph{International Conference on Machine Learning (ICML)}.

\bibitem[{Singh et~al.(2019)Singh, Natarajan, Shah, Jiang, Chen, Batra, Parikh, and Rohrbach}]{singh2019towards}
Singh, A.; Natarajan, V.; Shah, M.; Jiang, Y.; Chen, X.; Batra, D.; Parikh, D.; and Rohrbach, M. 2019.
\newblock Towards vqa models that can read.
\newblock In \emph{Proceedings of the IEEE/CVF conference on computer vision and pattern recognition}, 8317--8326.

\bibitem[{Team et~al.(2023)Team, Anil, Borgeaud, Wu, Alayrac, Yu, Soricut, Schalkwyk, Dai, Hauth et~al.}]{team2023gemini}
Team, G.; Anil, R.; Borgeaud, S.; Wu, Y.; Alayrac, J.-B.; Yu, J.; Soricut, R.; Schalkwyk, J.; Dai, A.~M.; Hauth, A.; et~al. 2023.
\newblock Gemini: a family of highly capable multimodal models.
\newblock \emph{arXiv preprint arXiv:2312.11805}.

\bibitem[{Touvron et~al.(2023)Touvron, Lavril, Izacard, Martinet, Lachaux, Lacroix, Rozi{\`e}re, Goyal, Hambro, Azhar et~al.}]{touvron2023llama}
Touvron, H.; Lavril, T.; Izacard, G.; Martinet, X.; Lachaux, M.-A.; Lacroix, T.; Rozi{\`e}re, B.; Goyal, N.; Hambro, E.; Azhar, F.; et~al. 2023.
\newblock Llama: Open and efficient foundation language models.
\newblock \emph{arXiv preprint arXiv:2302.13971}.

\bibitem[{Wang et~al.(2024)Wang, Chen, Zhou, Zhu, Liang, Shan, Liu, Xu, Yang, and Qin}]{wang2023smarttrim}
Wang, Z.; Chen, J.; Zhou, W.; Zhu, H.; Liang, J.; Shan, L.; Liu, M.; Xu, D.; Yang, Q.; and Qin, B. 2024.
\newblock {S}mart{T}rim: Adaptive Tokens and Attention Pruning for Efficient Vision-Language Models.
\newblock In \emph{Proceedings of the 2024 Joint International Conference on Computational Linguistics, Language Resources and Evaluation.}, 14937--14953.

\bibitem[{Xu et~al.(2023)Xu, Li, Chen, Chang, Liu, and Wang}]{xu2023no}
Xu, X.; Li, C.; Chen, Y.; Chang, X.; Liu, J.; and Wang, S. 2023.
\newblock No Token Left Behind: Efficient Vision Transformer via Dynamic Token Idling.
\newblock In \emph{Australasian Joint Conference on Artificial Intelligence}, 28--41. Springer.

\bibitem[{Yuan, Li, and Sun(2023)}]{yuan2023tinygpt}
Yuan, Z.; Li, Z.; and Sun, L. 2023.
\newblock Tinygpt-v: Efficient multimodal large language model via small backbones.
\newblock \emph{arXiv preprint arXiv:2312.16862}.

\bibitem[{Yuan et~al.(2024)Yuan, Shang, Zhou, Dong, Xue, Wu, Li, Gu, Lee, Yan et~al.}]{yuan2024llm}
Yuan, Z.; Shang, Y.; Zhou, Y.; Dong, Z.; Xue, C.; Wu, B.; Li, Z.; Gu, Q.; Lee, Y.~J.; Yan, Y.; et~al. 2024.
\newblock LLM Inference Unveiled: Survey and Roofline Model Insights.
\newblock \emph{arXiv preprint arXiv:2402.16363}.

\end{thebibliography}
\section{Appendix}
\subsection{Visualization}
In order to visually demonstrate the effectiveness of our proposed method, we have added additional visualization experiments. The red box area represents the image area corresponding to the answer.

\begin{figure}[!htb]
\centering
\includegraphics[width=1.\linewidth]{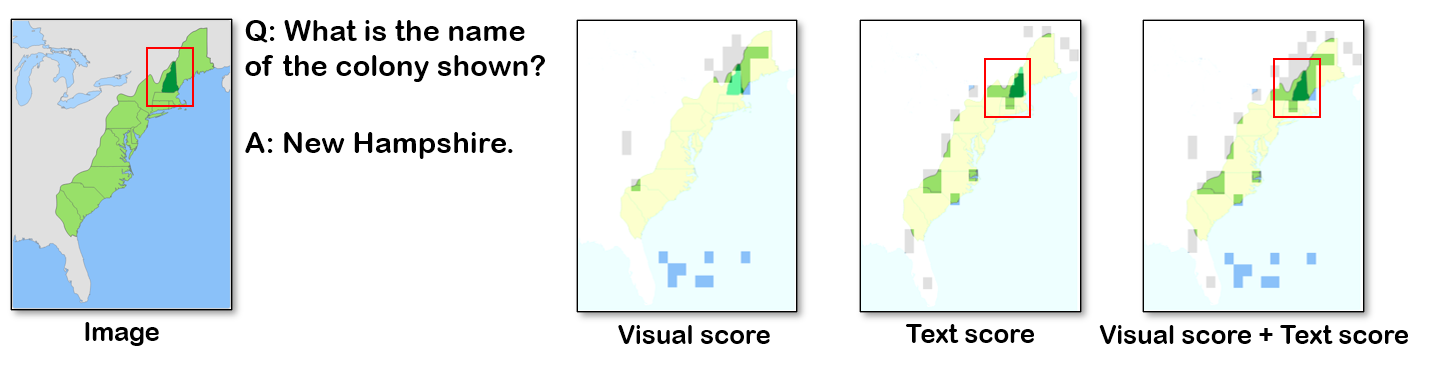}
\caption{Visualization results of token select/recovery with visual/text scores.} 
\label{fig: v2}
\end{figure}

As shown in Figure \ref{fig: v2}, in this instance, the visual score has already selected some tokens related to the question, and the tokens obtained using the text information recovery mechanism further increase the tokens associated with the problem.

\begin{figure}[!htb]
\centering
\includegraphics[width=1.\linewidth]{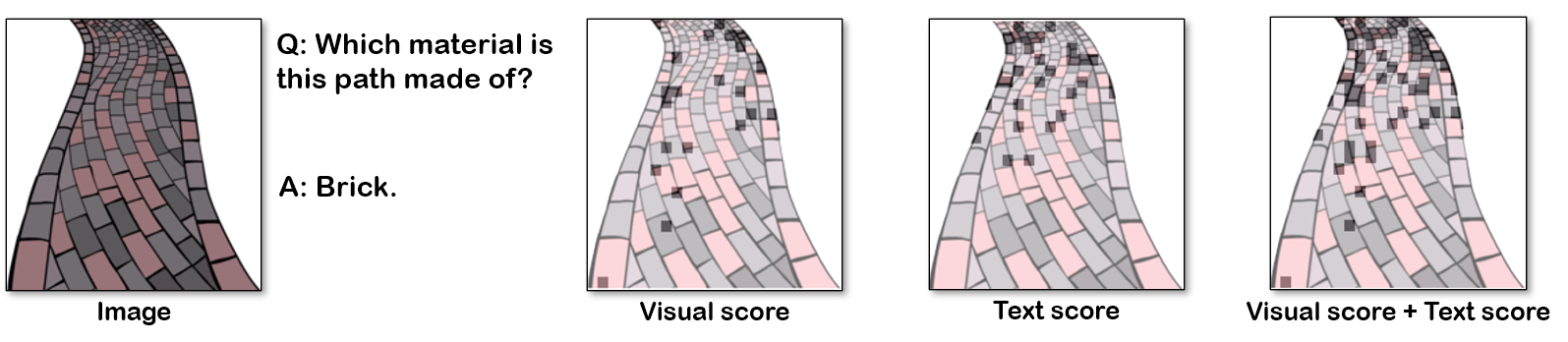}
\caption{Visualization results of token select/recovery with visual/text scores.} 
\label{fig: v3}
\end{figure}  

\begin{figure}[!htb]
\centering
\includegraphics[width=1.\linewidth]{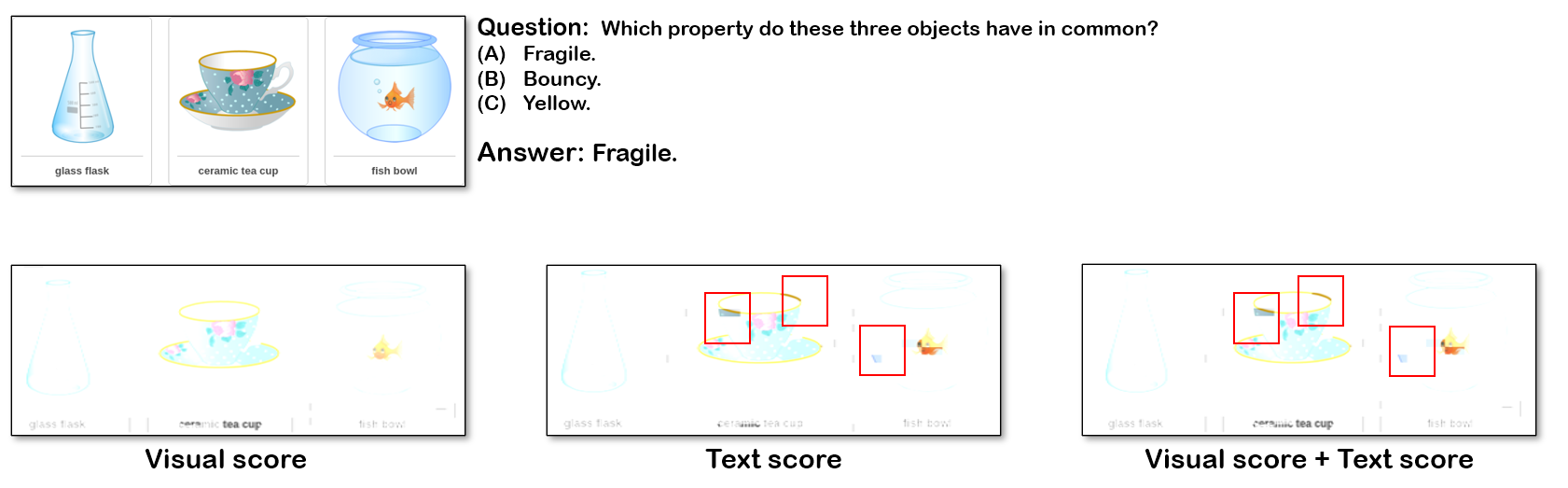}
\caption{Visualization results of token select/recovery with visual/text scores.} 
\label{fig: v4}
\end{figure}  

In Figure \ref{fig: v3}, The visual score has selected tokens related to the question area, and the text information recovery mechanism continues to supplement tokens related to the problem to ensure the model.

Figure \ref{fig: v4} shows a summary example of a question that requires the model to select the best option. The areas with high visual scores are mostly concentrated in the text area, but these areas are not highly relevant to the question. The token obtained by the text information recovery mechanism focuses on the edge position of the entity region in the image. And it happens to correspond to the fragility of the attribute, which helps the model choose the most general and correct option. However, for the entity beaker, neither the tokens selected by the visual score nor the text information recovery mechanism have been paid attention to.

\subsection{OCR-related Benchmark Results}
\begin{table}[!htb]
\centering
\caption{Performance on the OCR-related Benchmark}
\label{tab: ocr}
\begin{tabular}{lll}
\hline
Method   & DocVQA       & ChartQA       \\ \hline
LLaVA1.5 & 22.7 (100\%)       & 17.8 (100\%)         \\ \hline
Ours     & 16.7 (9.8\%) & 16.29 (9.9\%) \\ \hline
\end{tabular}%
\end{table}
To offer a more comprehensive understanding of the model's capabilities across different scenarios. We compared our method with the base model LLaVA-1.5-7B under the training-free setting, and the experimental results are shown in Table\ref{tab: ocr}. Our method still has good performance in the OCR realized benchmark, with only a slight decrease in performance while significantly reducing computational consumption.
\end{document}